\documentclass[journal]{IEEEtran}
\usepackage[utf8]{inputenc}
\usepackage{cite}
\usepackage{amsmath}
\usepackage{amssymb}
\usepackage{amsfonts}
\usepackage{algorithmic}
\usepackage{algorithm}
\usepackage{xcolor}
\usepackage{wrapfig, graphicx}
\usepackage{mathtools}
\usepackage{subfig}

\DeclarePairedDelimiter\floor{\lfloor}{\rfloor}
\newtheorem{assumption}{Assumption}
\newtheorem{theorem}{Theorem}

\newtheorem{remark}{Remark}

\begin{document}
%
\title{Differential Privacy Meets Federated Learning under Communication Constraints}

\author{\IEEEauthorblockN{
    Nima Mohammadi,
    Jianan Bai,
    Qiang Fan,
    Yifei Song,
    Yang Yi, and
    Lingjia Liu
}

\thanks{The authors are with the Bradley Department of Electrical and Computer Engineering, Virginia Tech, Blacksburg, VA 24060 USA. The corresponding author is L. Liu (ljliu@ieee.org).
}}

%

\maketitle

\begin{abstract}
The performance of federated learning systems is bottlenecked by communication costs and training variance. The communication overhead problem is usually addressed by three communication-reduction techniques, namely, model compression, partial device participation, and periodic aggregation, at the cost of increased training variance. Different from traditional distributed learning systems, federated learning suffers from data heterogeneity (since the devices sample their data from possibly different distributions), which induces additional variance among devices during training. Various variance-reduced training algorithms have been introduced to combat the effects of data heterogeneity, while they usually cost additional communication resources to deliver necessary control information. Additionally, data privacy remains a critical issue in FL, and thus there have been attempts at bringing Differential Privacy to this framework as a mediator between utility and privacy requirements. This paper investigates the trade-offs between communication costs and training variance under a resource-constrained federated system theoretically and experimentally, and  how communication reduction techniques interplay in a differentially private setting. The results provide important insights into designing practical privacy-aware federated learning systems.
\end{abstract}

\begin{IEEEkeywords}
Federated Learning, Differential Privacy, Artificial Intelligence, Communication Constraints, and Training Variance
\end{IEEEkeywords}

\IEEEpeerreviewmaketitle

\section{Introduction}
\IEEEPARstart{W}{ith} the increasing importance of data privacy, federated learning emerges as a promising machine learning framework that enables the training of a shared model among multiple end devices and a parameter server without exchanging local data~\cite{mcmahan2016communication, smith2017federated, yang2019federated, bagdasaryan2018backdoor}. 
Assuming a total of $N$ local devices and each device $i\in\{1,2,\cdots,N \}$ possesses a local dataset $\mathcal{D}_i$ with $|\mathcal{D}_i|$ samples, the local objective of device $i$ can be formulated as the following risk minimization problem

\begin{equation}
\min_\mathbf{x} f_i(\mathbf{x}) = \mathbb{E}_{\xi\sim\mathcal{D}_i} \ell(\mathbf{x};\xi),
\end{equation}

where $\mathbf{x}\in\mathbb{R}^d$ denotes the model parameter, $\xi\in\mathbb{R}^u$ is a training sample, and $\ell: \mathbb{R}^d \times \mathbb{R}^u \rightarrow \mathbb{R}$ is the sample-wise loss function. 
On the other hand, the goal of the parameter server is to find a single global model that would work well on the whole dataset $\mathcal{D} = \mathcal{D}_1\cup\mathcal{D}_2\cup\cdots\cup\mathcal{D}_N$. 
Accordingly, the global objective function can be represented as $f(\mathbf{x}) = \sum_{i=1}^Nw_if_i(\mathbf{x})$, where $w_i = {|\mathcal{D}_i|}/{|\mathcal{D}|}$ is the device weight proportional to the sample size~\cite{konevcny2016federated}.

Despite the privacy advantages of federated learning, which, as we discuss later, are prone to some challenges, its real-world implementation raises some issues regarding communication overhead and training variance:

\subsection{Communication Overhead}

Since the parameter server has no access to the local datasets, it needs to collect the local model updates from the local devices periodically, then send the aggregated model back to all devices. 
The communication rounds between the parameter server and the local devices result in a substantial communication overhead, especially when the number of devices is large. Moreover, the underlying communication between the parameter server and the local devices are usually imperfect and with limited capacity, and hence, it is imperative to design reliable and efficient federated learning algorithms under communication budget constraints.

From a communication point of view, federated learning systems realize model aggregation (uplink) through the multiple access channel (MAC)~\cite{wyner1994shannon}, and model distribution (downlink) through the broadcast channel (BC)~\cite{caire2003achievable}. 
Since a substantial number of local devices will send potentially different local model updates to the parameter server during model aggregation, the limited capacity of the uplink communications is usually the bottleneck of the system. 
Therefore, the existing literature suggests the following communication-reduction strategies to be utilized for enhancing the communication efficiency of federated learning systems, especially for the uplink (MAC) part:

\subsubsection{Model Compression} The transmission of each full-accuracy single-precision floating-point value requires $32$ bits. 
Since the model size is generally large in machine learning systems, transmitting model parameters with full accuracy can be prohibitively expensive. 
In this regard, a common strategy is to quantize the local model updates with some low-accuracy compressors or send only some important local parameters. 
Overall, the communication overhead can be significantly reduced via compression.

\subsubsection{Partial Participation} Due to the straggler's effect (some devices becoming non-responding and inactive), the response time for some devices can be prohibitively long. 
Therefore, awaiting model updates from all devices is not desirable in practice. 
Furthermore, the capacity of the underlying MAC channel is usually limited and does not linearly increase with the number of transmitting devices.
A natural solution to resolve these issues is partial participation through scheduling: only $M < N$ devices will be scheduled to transmit during each communication slot (round). 

\subsubsection{Periodic Aggregation} The aggregation of local models requires synchronization among all active devices. 
Aggregating in each iteration of training, as in traditional distributed learning, results in a large communication overhead. 
Therefore, frequent model synchronization may become unrealistic and consume a lot of system resources for communication.
A widely adopted strategy is to conduct a number of local iterations before synchronizing with the parameter server to save communication overhead.
However, since all devices perform local updates in an unsynchronized way, the update direction can deviate from global gradient direction especially for non-i.i.d. settings.

\looseness=-1
It is important to note that the three communication-reduction strategies are inherently coupled under communication constraints.
For example, the payload of the model parameters will inevitably impact how many devices ($M$) can be active under a fixed capacity of the MAC channel. 
Meanwhile, there is a clear trade-off between the payload of the model parameters and the period of the model aggregation under a fixed MAC capacity constraint: larger payload will lead to a less frequent model aggregation.
Therefore, a joint analysis seems necessary to provide a comprehensive analysis of federated learning systems under communication constraints. 

\subsection{Training Variance}
The training variance in federated learning can come from different sources. A universal one is the variance of the stochastic gradient, which exists in all stochastic gradient decent (SGD)-based training algorithms. This variance is induced because in SGD-based training algorithms, instead of evaluating the true gradient $\nabla f(\mathbf{x})$ computed over the whole training dataset, which costs expensive computing resources, an estimation $\tilde{\nabla} f(\mathbf{x})$ is computed only over a mini-batch of the training dataset. Although $\tilde{\nabla} f(\mathbf{x})$ is usually an unbiased estimate of $\nabla f(\mathbf{x})$~\cite{bottou2010large}, it has variance, given by $\mathbb{E}\|\tilde{\nabla} f(\mathbf{x}) - \nabla f(\mathbf{x}) \|^2$, that depends on the size of the mini-batches. Compared to centralized SGD that only has the stochastic gradient variance, federated learning systems suffer from the variance induced by imperfect communication and data heterogeneity.

\subsubsection{Data Heterogeneity}
In practice, the local datasets $\mathcal{D}_i$, $i \in \{1,\cdots,N\}$, are drawn from possibly different and unknown distributions $\mathcal{P}_i$, $i \in \{1,\cdots,N\}$. Thus, the local objective functions, $f_i(\mathbf{x})$'s, are non-uniform (different) among the local devices and can deviate from the global objective $f(\mathbf{x})$. 
Thus, the presence 
of data heterogeneity renders the training and analysis of federated learning systems more challenging compared with traditional distributed learning systems.

\subsubsection{Imperfect Communication}
To see how imperfect communication results in additional training variance, we assume there is no stochastic gradient variance, i.e., $\tilde{\nabla}f(\mathbf{x}) = {\nabla}f(\mathbf{x})$, and examine the effects of the three communication-reduction techniques individually. First, when model compression is used, the gradient direction will be quantized as $Q(\nabla f(\mathbf{x}))$, which can deviate from $\nabla f(\mathbf{x})$. Although some stochastic compressors can provide an unbiased estimate of  $\nabla f(\mathbf{x})$, the variance cannot be eliminated. Second, when the set of participating devices, $\mathcal{S}$, does not contain all the local devices, the aggregated gradient $\sum_{i\in\mathcal{S}}\nabla f_i(\mathbf{x})$ may not align with $\nabla f(\mathbf{x})$. Third, under periodic aggregation, the devices can generate different local models that result in client drift~\cite{scaffold}.

\subsubsection{Variance Reduction}
To combat the training variance among local devices, many variance-reduction techniques are developed for federated learning. Some of them originate from DANE~\cite{shamir2014communication}, which is a classical optimization method that introduces a sequence of local subproblems to reduce client drift. Fed-DANE~\cite{li2019feddane} is adapted from DANE by allowing partial device participation. Network-DANE~\cite{li2020communication} is developed for decentralized federated learning. SCAFFOLD~\cite{scaffold} can also be viewed as an improved version of DANE in federated settings by introducing some control variates.

\subsection{Local Differential Privacy}
Federated Learning achieves some levels of privacy by keeping the local datasets on user devices and only sharing the local updates with the server. This, however, has been proven to be insufficient for maintaining data privacy as the parameters can reveal insights into the data that has been used for training. Consequently, FL by itself can only be incorporated with honest participating parties, and to extend it for secure and privacy-preserving settings, extra measures should be considered.

By design, federated learning is ignorant of how the local updates are being generated, making it vulnerable to different forms of poisoning attacks from one or more malicious users \cite{bagdasaryan2020backdoor}.  Also, sharing the raw gradients can impose privacy risks for clients that can be exploited by a curious aggregation server, an adversary eavesdropping on the transmitted local updates, or a malicious participating client who might or might not be aware of the architecture of the model. A consequence of this is membership inference attack \cite{shokri2017membership} which allows an adversary with access to a number of records to infer whether a specific record was part of the training data (i.e., local private data of one of the clients) or not. 

\looseness=-1
With the lack of a rigorous privacy guarantee for FL, there have been attempts to bring the de facto framework of privacy-preserving analysis, Differential Privacy (DP) \cite{dwork2008differential}, and its local counterpart, Local DP \cite{duchi2013local}, to Federated Learning. 
Privatizing machine learning by adding noise has introduced different methods based on the stage noise addition takes place. Specifically, for an Empirical Risk Minimization problem, three main approaches have been introduced for differentially private optimization: 1) \textit{Objective Perturbation} where a randomized regularization term is added to the loss function, 2) \textit{Output Perturbation} where noise is added to the parameters of a non-private model after training, and 3) the currently prevailing \textit{Gradient Perturbation}, a more practical method that adds noise to the released gradient at each step, drawing more attention as it has been shown to be effective for nonconvex problems (as opposed to the last two methods) \cite{chaudhuri2011differentially,Yu2018ImproveTG}. 

To achieve data privacy in FL, instead of submitting raw local updates, the local parameters can be first perturbed using a randomization algorithm and then be released to the parameter server (local model), or alternatively, in the centralized fashion, the trusted parameter server can add the noise to the aggregated updates (curator model). This addition of noise ensures that the local updates remain private and do not leak unnecessary information. Notice that there is a clear trade-off between the utility (accuracy of the model that is being trained) and the preserved privacy achieved by the noise. The perturbation is to impede attempts to infer the true values of a client with strong confidence but still allow accurate inferences for the population.

In the DP setting, it is assumed that the party responsible for aggregating the results, the parameter server, is trusted. Therefore the privacy could be maintained by adding noise to the aggregated results. However, for the surging edge computing and IoT applications, the parameter server ideally should not be trusted. This necessity naturally drives the research toward the local mode of DP, where each client would perturb its data to ascertain the data is kept private. This is to ensure that the clients' privacy is maintained even from the aggregator or an attacker that gets access to the data of the client on the server. However, the downside of the local mode is that the accumulated noise would incur more accuracy loss compared to the centralized context, mandating the need for more train data and longer training.

\looseness=-1
The privacy parameters $(\varepsilon, \delta)$ quantify DP where for smaller values thereof we get more privacy. Formally, a randomized algorithm $\mathcal{A}$ is $(\varepsilon, \delta)$ -differentially private if for all $\mathcal{S} \subseteq \operatorname{Range}(\mathcal{A}),$ and for all adjacent datasets $D$ and $D^{\prime}$, then we have:
\begin{equation}
    \operatorname{Pr}[\mathcal{A}(D) \in \mathcal{S}] \leq e^{\varepsilon} \operatorname{Pr}\left[\mathcal{A}\left(D^{\prime}\right) \in \mathcal{S}\right]+\delta
\end{equation}

where $D$ and $D^{\prime}$ are two datasets that only differ in a single entry. In this context, the randomized algorithm $\mathcal{A}$ provides privacy by making the two datasets difficult to distinguish.. However, in absence of trust with the data collector, local DP is deemed more suitable. Formally, a randomized mechanism $\mathcal{M}$ is $(\varepsilon, \delta)$-LDP, for any pair of inputs $x$ and $x^{\prime}$ in $\mathcal{X}$, and any measurable subset $\mathcal{O} \subseteq \operatorname{Range}(\mathcal{M})$, then we have:
\begin{equation}
    \operatorname{Pr}[\mathcal{M}(x) \in \mathcal{O}] \leq e^{\epsilon} \cdot \operatorname{Pr}\left[\mathcal{M}\left(x^{\prime}\right) \in \mathcal{O}\right]+\delta
\end{equation}
Again, the privacy guarantee of $\mathcal{M}$ is determined by $\varepsilon$, but with low probability of $\delta$ this might not hold.


\subsection{Contributions}

Although the effects of model compression, partial device participation, periodic aggregation, and data heterogeneity have been studied, a few works in the literature provide a comprehensive analysis by jointly considering all of them. This paper, to the best of our knowledge, is the first one presenting the convergence analysis of stochastic gradient descent in privacy-preserving federated settings by considering all the three communication-reduction techniques under the presence of data heterogeneity for strongly convex loss functions. Based on the results, we can have a clear understanding of how these components affect the convergence of federated learning systems and interplay with each other. We investigate differential privacy in the context of federated learning by introducing privacy-augmented FedPaq and discussing the impact of parameters describing the gradient perturbation on performance of the model. Moreover, a privacy amplification method, based on subsampling of local datasets, is employed to enhance the convergence rate of the privatized model.
In addition, our analysis provides practical insights into the design of a privatized FL model and communication strategies to accelerate the training process of such systems under a limited communication budget.
To be specific, since the all the parameters involved in the privacy measure and the three communication-reduction techniques are jointly considered, we are able to quantitatively analyze the trade-offs between different options. As a result, important design intuitions for real-world differentially private federated learning systems that are limited by the communication capacity constraints in wireless networks are provided.

 
\subsection{Related Works}

There have been several works on the convergence analysis of federated learning systems with different communication-reduction approaches.

The convergence rate of the FedAvg algorithm, which was first introduced in \cite{mcmahan2016communication}, has been studied in a non-i.i.d. data setting with partial device participation and periodic aggregation in~\cite{fedavg}. However, the authors did not consider quantization. Similar set of results were presented in~\cite{local_gd}. In~\cite{fedpaq}, the 
FedPaq framework was introduced and analyzed under all three communication-reduction approaches. However, the authors assumed i.i.d. (homogeneous) data. To combat the effect of data heterogeneity, 
~\cite{fedprox1, fedprox2} introduced an adaptation of FedAvg, named FedProx. In FedProx, a proximal term is introduced to each local objective function. However, these two works did not consider quantization.
Finally, and perhaps the most recent work,~\cite{scaffold} introduced the SCAFFOLD framework that incorporates a variance-reduction mechanism to combat the effect of non-i.i.d. data. More specifically, in this  framework, some control variates are introduced to reduce the drifts among different devices. While in~\cite{scaffold}, the authors show SCAFFOLD achieves better performance compared with FedAvg, the performance improvement is not significant under moderate heterogeneity (e.g., $10\%$ similarity). On the other hand, since in SCAFFOLD all active devices need to send an additional control variate update, which has the same dimension as the model parameter, to the parameter server, the communication cost is doubled in each round. In practice, FedPaq is still a good candidate federated learning algorithm. However, in the original paper~\cite{fedpaq}, the convergence analysis relies on the i.i.d. data assumption, which is unrealistic. To investigate the trade-offs between communication costs and training variance, it is important to obtain the convergence results for FedPaq under data heterogeneity, which is one of the main contributions of this paper.

In the literature, various randomized mechanisms and variations of DP and LDP, referred to as protocols, have been investigated to bring privacy guarantees to FL. In \cite{gursoy2019secure}, the notion of Condensed LDP ($\alpha-$CLDP) is introduced which is later used in \cite{sun2020ldp} to propose LDP-FL. Developing quantization schemes are more compatible with continious Gaussian and Laplacian noise is a recent theme of the research. Adding continuous noise after quantization, would turn the value into a continuous number and the benefits of compression is lost. For LDP, biased compressors can not be used as they break the independence between rounds. This has led to works on discrete noise addition and compressors that take privacy perturbation into account. cpSGD overcomes the prohibitive problem of discrete values by adding noise drawn from a Binomial distribution and showing that for small $\varepsilon$ it mimics the Gaussian mechanism \cite{agarwal2018cpsgd}.
In \cite{canonne2020discrete}, a discrete Gaussian noise is introduced which has been used in \cite{wang2020d2pfed}

\begin{figure*}[htb!]
    \centering
    \includegraphics[width=12cm]{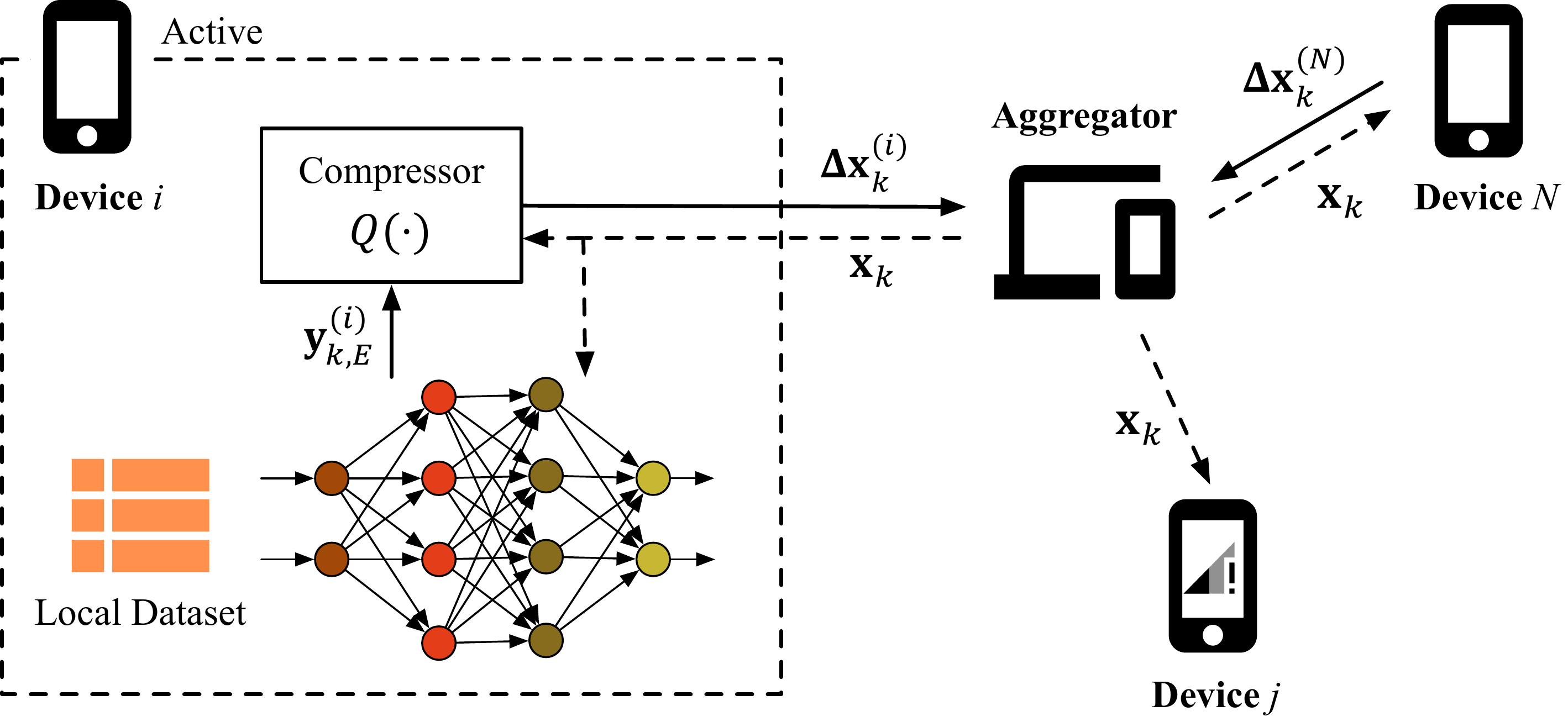}
    \caption{Federated learning system.}
    \label{fig: system}
\end{figure*}

\section{Differential Privacy Meets Federated Learning under Communication Constraints}

We start by introducing a general framework of federated learning system, which can be seen in Fig.~\ref{fig: system}. The system consists of $N$ local devices located in a wireless network and they are all connected to a parameter server, which coordinates all devices to train a shared model. In a given communication round $k\in[K]$, the parameter server first distributes the aggregated information $(\mathbf{x}_k, \mathbf{c}_k)$, which is obtained through last communication round, to all active devices, where $\mathbf{x}_k$ is the global model and $\mathbf{c}_k$ is the global control variate\footnote{We use $[K]$ to represent the set $\{1,\cdots,K\}$.}. 
On the other hand, for a device $i$ that belongs to the set of participating (active) devices, $\mathcal{S}_k$, it will calculate the local model update $\Delta\mathbf{x}_k^{(i)} := \mathcal{V}_i(\mathbf{x}_k, \mathbf{c}_k; \mathcal{D}_i)$ and the update of the control variate $\Delta \mathbf{c}_k^{(i)}$, for some local functions $\mathcal{V}_i , i \in \mathcal{S}_k$. The complete training process is summarized in Algorithm~\ref{alg: general alg}.

\begin{algorithm}
\caption{General Federated Learning}
\label{alg: general alg}
\begin{algorithmic}[1]
 \renewcommand{\algorithmicrequire}{\textbf{Input:}}
 \renewcommand{\algorithmicensure}{\textbf{Initialize:}}
 \newcommand{\algorithmicinitialize}{\textbf{Initialize:}}
 \REQUIRE global learning rate $\eta_{k,g}$ for $k\in[K]$
 \ENSURE model parameters $\mathbf{x}_0 \in \mathbb{R}^d$
  \FOR {each round $k = 0,\cdots,K-1$}
  \STATE \textbf{on each device} $i\in\mathcal{S}_k$:
  \begin{ALC@g}
  \STATE calculate $\Delta\mathbf{x}_k^{(i)} = \mathcal{V}_i(\mathbf{x}_k, \mathbf{c}_k; \mathcal{D}_i)$
  \STATE (optional) calculate $\Delta \mathbf{c}_k^{(i)}$
  \STATE send $\Delta \mathbf{x}_k^{(i)}$ and $\Delta\mathbf{c}_k^{(i)}$ to the parameter server
  \end{ALC@g}
  \STATE \textbf{on the parameter server}:
  \begin{ALC@g}
  \STATE collect the local updates from devices in $\mathcal{S}_k$
  \STATE calculate $\mathbf{x}_{k+1} = \mathbf{x}_k + \frac{\eta_{k,g}}{M}\sum_{i\in\mathcal{S}_k}\Delta\mathbf{x}_k^{(i)}$
  \STATE (optional) calculate $\mathbf{c}_{k+1}$
  \STATE broadcast $\mathbf{x}_{k+1}$ and $\mathbf{c}_{k+1}$ to all devices.
  \end{ALC@g}
  \ENDFOR
\end{algorithmic} 
\end{algorithm}

\subsection{Federated Learning Algorithms}
Different designs of $\mathcal{V}_i$ and $\mathcal{S}_k$ lead to different algorithms. Here, we introduce some exemplary algorithms:
\subsubsection{Distributed SGD}
In distributed SGD~\cite{chen2016revisiting}, all devices participate in each communication round, i.e., $\mathcal{S}_k = [N]$ for all $k\in[K]$. Additionally, the local model update is calculated by one local SGD step, such that 
\begin{equation}
    \Delta \mathbf{x}_k^{(i)} = -\eta_{k,l}\tilde{\nabla} f_i(\mathbf{x}_k).
\end{equation}
where $\eta_{k,l}$ is the local learning rate used at communication round $k$. Distributed SGD does not include a control variate. However, since it requires full participation and communication happens in each training round, it can result in overwhelming communication costs.

\subsubsection{FedAvg and FedPaq} 
FedAvg~\cite{fedavg} is proposed to save communication costs by using partial participation and periodic aggregation. Let $E$ be the number of local iterations, then a device $i\in\mathcal{S}_k$ generates a sequence of local models $\left\{\mathbf{y}_{k,0}^{(i)}, \mathbf{y}_{k,1}^{(i)}, \cdots, \mathbf{y}_{k,E}^{(i)} \right\}$ with
\begin{equation}
\label{eq: FedAvg1}
\mathbf{y}_{k,0}^{(i)} = \mathbf{x}_k\ \ \text{and}\ \ \mathbf{y}_{k,t+1}^{(i)} = \mathbf{y}_{k,t}^{(i)} - \eta_k^l\tilde{\nabla} f_i\left(\mathbf{y}_{k,t}^{(i)} \right),
\end{equation}
for $t\in[E]$. 
Then, the local model update is given by
\begin{equation}
\label{eq: FedAvg2}
    \Delta \mathbf{x}_k^{(i)} = \mathbf{y}_{k,E}^{(i)} - \mathbf{x}_k.
\end{equation}
Similarly, FedPaq~\cite{fedpaq} also generates the sequence of local models in \eqref{eq: FedAvg1}, but clients would send quantized version of local updates to the parameter server, i.e.,
\begin{equation}
    \Delta\mathbf{x}_k^{(i)} = Q\left(\mathbf{y}_{k,E}^{(i)} - \mathbf{x}_k \right)
\end{equation}
to further reduce communication overhead.  Notice that FedPaq is a generalization of both distributed SGD (with $E=1$, $\mathcal{S}_k=[N]$, and $Q(\mathbf{x})=\mathbf{x}$) and FedAvg (with $Q(\mathbf{x})=\mathbf{x}$). 

\subsubsection{SCAFFOLD} 
Instead of updating the local models only by the local stochastic gradients $\nabla f_i(\mathbf{y})$, SCAFFOLD~\cite{scaffold} uses some control variates to reduce the drifts that occur among different devices. To be specific, one local step in SCAFFOLD is given by
\begin{equation}
    \mathbf{y}_{k,t+1}^{(i)} = \mathbf{y}_{k,t}^{(i)} - \eta_k^l\left(\tilde{\nabla}f_i\left(\mathbf{y}_{k,t}^{(i)} \right) - \mathbf{c}_{k}^{(i)} + \mathbf{c}_k \right),
\end{equation}
while the local control variate $\mathbf{c}_k^{(i)}$ and the global control variate $\mathbf{c}_k$ are updated by
\begin{equation}
    \mathbf{c}_{k+1}^{(i)} = \mathbf{c}_{k}^{(i)} - \mathbf{c}_k - \frac{1}{E\eta_k^l}\Delta\mathbf{x}_k^{(i)},
\end{equation}
\begin{equation}
    \mathbf{c}_{k+1} = \mathbf{c}_k + \frac{1}{N}\sum_{i\in\mathcal{S}_k}\left(\mathbf{c}_{k+1}^{(i)} - \mathbf{c}_{k}^{(i)} \right).
\end{equation}
The idea behind SCAFFOLD is to mimic the ideal update under centralized SGD, i.e.,
\begin{equation}
    \tilde{\nabla}f_i\left(\mathbf{y}_{k,t}^{(i)} \right) - \mathbf{c}_{k}^{(i)} + \mathbf{c}_k \approx \frac{1}{N}\sum_{j\in[N]}\tilde{\nabla}f_j\left(\mathbf{y}_{k,t}^{(j)} \right).
\end{equation}

\subsection{Communication Constraints}

To analyze the trade-offs between communication costs and training variance, we focus our analysis on FedPaq, which jointly considers model compression, partial device participation, and periodic aggregation under data heterogeneity.  To simplify the analysis while preserving the essence of the communication constraint, we assume the capacity of the underlying $M$-user MAC channel for the federated learning system is bounded by $\mathcal{C}$ bits/second~\cite{cover2012elements} and the total duration of the training process is $\mathcal{T}$ seconds. 
During the training process, each local device can conduct a total of $T$ training iterations while the total of $B = \mathcal{C}  \mathcal{T}$ bits can be shared among the active devices participating the model aggregation process. 
Accordingly, we can link the communication constraints of the federated learning system of interests in the following:
\begin{equation}
    B = \mathcal{C}  \mathcal{T} = K M \beta = \floor*{\frac{T}{E}} M \beta \approx \frac{TM\beta}{E},
    \label{equ_CommCon}
\end{equation}
where $K = \left\lfloor \frac{T}{E} \right\rfloor$ is the total number of training rounds and $\beta$ is the number of bits required to transmit the model update.
In this way, we can provide a unified framework to conduct performance analysis of federated learning under communication constraints.

\begin{algorithm}
\caption{Privacy Augmented FedPaq.}
\label{alg: SecFedPaq}
\begin{algorithmic}[1]
 \renewcommand{\algorithmicrequire}{\textbf{Input:}}
 \renewcommand{\algorithmicensure}{\textbf{Initialize:}}
 \newcommand{\algorithmicinitialize}{\textbf{Initialize:}}
 \REQUIRE $\eta_k$ for $k\in[K]$
 \ENSURE model parameters $\mathbf{x}_0 \in \mathbb{R}^d$
  \FOR {each round $k = [K]$}
  \STATE \textbf{on each device} $i\in\mathcal{S}_k$:
  \begin{ALC@g}
  \STATE Initialize the local model $\mathbf{x}_{k,0}^{(i)} = \mathbf{x}_k$
  \STATE Select a random subset $\mathcal{D}_{s}$ of size $Eb$ from $\mathcal{D}_{i}$
  \FOR {each iteration $t = 0,\cdots,E-1$}
  \STATE calculate $\mathbf{x}_{k, t+1}^{(i)} = \mathbf{x}_{k, t}^{(i)} - \eta_k \tilde{\nabla} f_i\left(\mathbf{x}_{k,t}^{(i)} \right)$
  \ENDFOR
  \STATE $\mathbf{z}_{k}^{(i)} \sim \mathcal{N}\left(\mathbf{0}, \sigma_{i, k}^{2} \boldsymbol{1}_{d}\right)$ 
  \STATE send $\Delta\mathbf{x}_{k}^{(i)} = Q\left(\mathbf{x}_{k,E}^{(i)} - \mathbf{x}_k + \mathbf{z}_{k}^{(i)}\right)$ to the server
  \end{ALC@g}
  \STATE \textbf{on the parameter server}:
  \begin{ALC@g}
  \STATE calculate $\mathbf{x}_{k+1} = \mathbf{x}_k +  \frac{1}{M}\sum_{i\in\mathcal{S}_k}\Delta\mathbf{x}_{k}^{(i)}$ 
  \STATE broadcast the global model $\mathbf{x}_{k+1}$ to all devices
  \end{ALC@g}
  \ENDFOR
  
\end{algorithmic} 
\end{algorithm}


\subsection{Privacy Measure}
Algorithm~\ref{alg: SecFedPaq} outlines the introduced privacy-aware version of FedPaq. After $E$ local iterations, the clients generate the local updates which are to be sent to the parameter server. Prior to uploading the local model updates, they get perturbed by a random noise drawn from a Gaussian distribution to preserve the privacy level of desire, i.e., a well-known procedure referred to as the Gaussian mechanism for achieving $(\varepsilon, \delta)$-DP privacy guarantee. To locally differentially privatize a function $\mathrm{f}(X)$ subject to $(\varepsilon, \delta)$ we use

\begin{equation}
    M(X,\mathrm{f}, \sigma) \triangleq \mathrm{f}(X)+\mathcal{N}\left(0, \sigma^{2}\mathbf{I}\right),
\end{equation}

with 

\begin{equation}
    \epsilon=\frac{\Delta_{f}}{\sigma} \sqrt{2 \ln \frac{1.25}{\delta}},
\end{equation}

for any $\delta \in(0,1]$ where $\Delta_{f}$ bounds the $L_2$ sensitivity of $\mathrm{f}(X)$, that is

\begin{equation}
    \left\|f(x)-f\left(x^{\prime}\right)\right\|_{2} \leq \Delta_{f}, \forall x, x^{\prime} \in X
\end{equation}

To reduce the amount of noise required to be added to the local updates to achieve a certain privacy guarantee, one may use privacy amplification techniques. Here we employ subsampling into mini-batches to achieve this reduction. An algorithm $f$ would reach a better privacy guarantee when applied on a random subsample of the dataset instead of the full sample. This is intuitively due to the fact that data not included in the subsample would enjoy full privacy, hence the privacy being amplified. Applying the $(\epsilon, \delta)$-DP randomized mechanism $\mathcal{M}$ on the subsampled data achieves $(\epsilon', h(\delta))$-DP mechanism $\mathcal{M}^\mathcal{S}$ for $0\leq\epsilon'\leq\epsilon$ and a function $h$ to be determined based on the sampling procedure\footnote{Poisson subsampling, sampling without replacement and sampling with replacement are possible options that have been studied in the literature.} \cite{wang2019subsampled}. For subsampling $E$ mini-batches of size $b$ without replacement we achieve $\left(\log \left(1+\left(1-\left(1-b / n_{k}\right)^{E}\right)\left(e^{\epsilon}-1\right)\right), \gamma \delta\right)$-DP for $\Delta \mathbf{x}_k^{(i)}$ where $\gamma := Eb/n_k$.

Notice that a $(\epsilon, \delta)$-DP mechanism $\mathcal{M}$ is also $(\epsilon', \delta')$ for any $\epsilon'\geq\epsilon$ and any $\delta'\geq\delta$. Then, 

\begin{equation}
\begin{split}
\log \left(1+\left(1-\left(1-b / n_{k}\right)^{E}\right)\left(e^{\epsilon}-1\right)\right) \\
\leq \log \left(1+\gamma\left(e^{\epsilon}-1\right)\right) \leq \gamma\left(e^{\epsilon}-1\right) \leq 2\gamma\epsilon,
\end{split}
\end{equation}

hence the introduced method satisfying at least $(\epsilon, \delta)$-DP for $\mathbf{x}_{k,E}^{(i)}$ via adding reduced Gaussian noise of
\begin{equation}
\label{as: noise eq}
\begin{aligned}
\sigma_{i, k}^{2} &=\frac{2\left(\Delta_{f}\right)^{2} \ln (1.25 /(\delta / \gamma))}{(\epsilon / 2 \gamma)^{2}}.
\end{aligned}
\end{equation}

Following this procedure, having fixed the privacy parameters $\varepsilon$ and $\delta$, the algorithm achieves higher utility (in terms of model accuracy) and converges faster. Needless to say, clipping the local gradients to a small value to maintain lower global sensitivity, and subsampling, although result in noise of less magnitude, also may inversely impact the convergence in the non-secure setting. Therefore, careful convergence analysis is deemed necessary.  

\section{Convergence of Privatized FedPaq Under Heterogeneity}

\label{subsec_ConAna}
In this section, we present the convergence analysis for the Privatized FedPaq algorithm under non-i.i.d. data setting (Algorithm~\ref{alg: SecFedPaq}) with the following assumptions:
\vspace{.05in}
\begin{assumption}
\label{as: L-smooth}
The loss function $\ell$ is $L$-smooth, such that
$
    \|\nabla \ell(\mathbf{x}) - \nabla\ell(\mathbf{y}) \| \leq L\|\mathbf{x} - \mathbf{y} \|
$ for any $\mathbf{x}, \mathbf{y} \in \mathbb{R}^d$. 
Consequently, $f_i$ and $f$ are also $L$-smooth.
\end{assumption}
\vspace{.05in}
\begin{assumption}
\label{as: strong convexity}
The loss function $\ell$ is $\mu$-strongly convex, such that
$
    \|\nabla \ell(\mathbf{x}) - \nabla \ell(\mathbf{y}) \| \geq \mu\|\mathbf{x}-\mathbf{y} \|
$
for any $\mathbf{x}, \mathbf{y} \in \mathbb{R}^d$.
Consequently, $f_i$ and $f$ are also $\mu$-strongly convex.
\end{assumption}
\vspace{.05in}
\begin{assumption}
\label{as: stochastic gradient}
The stochastic gradient is unbiased and variance-bounded, such that for any $\mathbf{x}\in\mathbb{R}^d$ and the mini-batch samples $\xi$, we have
$
    \mathbb{E}\left[\nabla f_i\left(\mathbf{x};\xi \right)\right] = \nabla f_i\left(\mathbf{x}\right)$ and $\mathbb{E}\left\|\nabla f_i\left(\mathbf{x};\xi \right) - \nabla f_i(\mathbf{x}) \right\|^2 \leq \sigma^2/b$
for all $i=1,2,\cdots,N$. Here, $b$ is the mini-batch size.
\end{assumption}
\vspace{.05in}
\begin{assumption}
\label{as: compressor}
$Q(\cdot)$ is an unbiased and $q$-lossy random compressor, such that for any $\mathbf{x}\in\mathbb{R}^d$
$\mathbb{E}\left[Q(\mathbf{x})\right] = \mathbf{x}$ and  
$\mathbb{E}\|Q(\mathbf{x}) - \mathbf{x} \|^2 \leq q\|\mathbf{x} \|^2$.
\looseness=-1
Note that $q$ is a real value in $[0,1]$, which is determined by the resolution of the compressor, $\beta$. When $q=0$, we have $Q(\mathbf{x})=\mathbf{x}$ and there is no quantization. For $q=1$, no information is being sent during communications. 
\end{assumption}
\vspace{.05in}
\begin{remark}
    While Assumptions \ref{as: L-smooth}, \ref{as: strong convexity}, and \ref{as: stochastic gradient} are widely used in literature, the unbiasedness of the compressed model stated Assumption~\ref{as: compressor} does not always hold in practice. For example, the SignSGD~\cite{bernstein2018signsgda} uses biased quantizer which makes the analysis more complicated. For the tractability of our analysis, we use an unbiased compressor throughout this paper. Examples for unbiased quantizers include quantized-SGD and TernGrad~\cite{alistarh2017qsgd, wen2017terngrad}, which preserve the true values in expectation. Furthermore, for quantized-SGD, \cite{alistarh2017qsgd} shows that $q = \min\{n/s^2, \sqrt{n}/s \}$, where $n$ is the block-size and $s$ is the quantization level. For simplicity, we assume $n = d$.
\end{remark}
\vspace{.05in}
\begin{remark}
    While the assumption of strong convexity is restrictive, it facilitates our analysis since the convergence rate can be directly measured by $\| \mathbf{x}_k - \mathbf{x}^*\|^2$. Additionally, $\mu>0$ allows us to use the learning rate $\eta_k = \frac{4\mu^{-1}}{kE+4E}$ in Theorem \ref{th: convergence}.  Relaxing this assumption is part of our future work.
\end{remark}
\vspace{.05in}
\begin{remark}
    Notice that Assumption \ref{as: stochastic gradient} is general for all choices of batch size when calculating the stochastic gradients. A larger batch size will lead to a smaller $\sigma$.
\end{remark}
\vspace{.05in}
\begin{remark}
    Notice that Assumptions \ref{as: L-smooth}, \ref{as: strong convexity}, and \ref{as: stochastic gradient} are also used in \cite{fedavg}, which proves the convergence rate for FedAvg under non-iid data. However, our paper differs from \cite{fedavg} in that we consider privacy concerns and compression of model updates, making our system more realistic. However, this also makes the underlying analysis more challenging where the original results cannot be directly applied. Moreover, \cite{fedavg} has an additional assumption that the expected squared norm of stochastic gradients is uniformly bounded, i.e., $\mathbb{E}\left\|\nabla f_i\left(\mathbf{x}_{k,t}^{(i)};\xi_{k,t}^{(i)} \right) \right\|^2 \leq G^2$, where $G>0$, for all devices and all time steps. 
\end{remark}
\vspace{.05in}
\begin{assumption}
\label{as: heterogeneity}
    We assume that for all $\mathbf{x}\in\mathbb{R}^d$, there exist a $\lambda > 0$, such that
    \begin{equation}
        \frac{1}{N}\sum_{i\in[N]}\left\|\nabla f_i(\mathbf{x}) - \nabla f(\mathbf{x}) \right\|^2 \leq \lambda^2.
    \label{eq: hetg}
    \end{equation}
    The value of $\lambda$ quantifies the degree of data heterogeneity among different local devices. When $\lambda=0$, we have $\nabla f_i(\mathbf{x}) = \nabla f(\mathbf{x})$ for all $i=1, 2, \cdots, N$, and there is no heterogeneity. It can be seen from (\ref{eq: hetg}) that larger $\lambda$ implicates higher degree of heterogeneity.
\end{assumption}
\vspace{.05in}
\begin{remark}
    \looseness=-1
    Different papers usually have different assumptions on data heterogeneity. For example, \cite{fedavg} defines $\Gamma = f^* - \sum_{i=1}^Nw_if_i^*$ and \cite{local_gd} defines $\varsigma^2 = \frac{1}{N}\sum_{i=1}^N\nabla f_i(x^*)$ for quantifying the degree of non-i.i.d. Although our measure of heterogeneity is different from these two papers, we note that both $\Gamma$ and $\varsigma$ can still be expressed using $\lambda$ under strongly convex setting.  
\end{remark}
\vspace{.05in}
\begin{assumption}
\label{as: stochastic gradient bound}
The stochastic gradient is uniformly bounded, such that for any $\mathbf{x}\in\mathbb{R}^d$ and the mini-batch samples $\xi$, we have
$\mathbb{E}\left\|\nabla f_i\left(\mathbf{x};\xi \right)\right\|^2 \leq G^2$
for all $i=1,2,\cdots,N$.
\end{assumption}
\vspace{.05in}

For the simplicity of our analysis, we additionally assume the device weights, $w_i$'s, to be uniform among all devices. However, as suggested in \cite{fedavg}, this does not result in the loss of generalization of our results. Indeed, by replacing the local objective as $\Tilde{f}_i(\mathbf{x}) = w_iNf_i(\mathbf{x})$ and transforming the value of $L$, $\mu$, $\sigma$, and $\lambda$ accordingly, the global objective becomes the simple average of transformed local objectives, i.e., $f(\mathbf{x}) = \frac{1}{N}\sum_{i=1}^N \Tilde{f}_i(\mathbf{x})$. Moreover, we assume the same gradient perturbation is being performed by all clients. 

Using the assumptions and lemmas above, the convergence rate of Algorithm \ref{alg: SecFedPaq} for the secure federated learning setting can be characterized in the following theorem:

\begin{theorem}
\label{th: convergence}
    Training the secure federated learning system using Algorithm \ref{alg: SecFedPaq}, under Assumptions \ref{as: L-smooth}, \ref{as: strong convexity}, \ref{as: stochastic gradient}, \ref{as: compressor}, \ref{as: heterogeneity} and \ref{as: stochastic gradient bound} for $T$ iterations and setting the learning rate in the $k$-th round as $\eta_k = \frac{4\mu^{-1}}{kE + 4E}$, the aggregated global model $\mathbf{x}_K$, where $K = \left\lfloor \frac{T}{E} \right\rfloor$, satisfies
\begin{equation}
\label{eq: theorem 1}
\begin{aligned}
&\mathbb{E}\|\mathbf{x}_K - \mathbf{x}^* \|^2 \leq \frac{16E^2}{T^2}\mathbb{E}\|\mathbf{x}_{k_0} - \mathbf{x}^* \|^2\\
        &\quad+ \frac{16}{\mu^2}\left(\frac{2qG^2}{M}+\frac{qG^2}{N}\right)\frac{E}{T}\\
        &\quad+ \frac{16}{\mu^2}\left( \frac{4e\sigma^2}{bM} + \frac{3L\lambda^2}{\mu} + \frac{\sigma^2}{bN} \right)\frac{1}{T}\\
        &\quad+ \frac{128e\lambda^2}{\mu^2M}\cdot\frac{E-1}{T} + \frac{128G^2}{\mu^2}\frac{(E-1)^2}{T}\\
        &\quad+ \frac{4096dG^2b^2(1+q)\ln\left(\frac{1.25Eb}{|\mathcal{D}_i|\delta} \right)}{M|\mathcal{D}_i|^2\epsilon^2}\frac{E^3}{T}
    \end{aligned}
\end{equation}
\end{theorem}

\begin{remark}
    For vanilla version of the algorithm, without addition of noise required for privacy, similar to \cite{fedavg}, \cite{fedpaq}, and \cite{fedprox1}, our result suggests a convergence rate as $\mathcal{O}\left(\frac{E^2}{T}\right)$, which requires $E=o\left(\sqrt{T}\right)$. However, this term appears only when data heterogeneity exists, i.e., $\lambda>0$. When there is no data heterogeneity among local devices, the training can get significantly accelerated.
\end{remark}
\vspace{.05in}


\begin{remark}
\looseness=-1
    Convergence gets slowed down when $q$, $E$ and $\lambda$ become larger, whereas larger $M$ (more active devices) can facilitate the convergence. However, since $\mathcal{O}\left(\frac{1}{M}\right)$ does not appear in all terms, the training does not enjoy a linear speedup.
\end{remark}
\vspace{.05in}

\begin{remark}
\looseness=-1
The convergence rate is also impacted by other factors such as $\epsilon$, $\delta$ and $G$. If $\epsilon$ and $\delta$ are small, the model convergence will slow down, thus necessitating more global iterations. Meanwhile, if $G$ increases, the right part of (\ref{eq: theorem 1}) will also increases accordingly, and thus delays the model convergence.

\end{remark}
\vspace{.05in}

\subsection{Impacts of Communication Constraints}

As shown in~\eqref{equ_CommCon}, various communication overhead reduction strategies will be coupled together under the constraint of MAC capacity. 
Therefore, it is important for us to characterize the impacts on the various trade-offs among model compression, partial participation, and periodic aggregation based on the conducted convergence analysis to obtain new system design intuitions for federated learning systems under communication constraints and differential privacy.

\subsubsection{Partial Participation v.s. Periodic Aggregation} Assuming the model compression strategy is fixed, we can write $M = \alpha E$, where $\alpha = {B}/{T\beta}$. Furthermore, we assume a large-scale federated learning system where $N\gg M$, such that the approximation of
$\frac{(N-M)(N-1)}{MN^2} \rightarrow \frac{1}{M}$ holds. 
When the number of total iterations $T$ is large enough, the dominating term in \eqref{eq: theorem 1} becomes
\begin{equation}
\label{eq: tradeoff 1}
    \mathcal{O}\left(\frac{aE^2 + b\lambda^2 + c\lambda^2/E + d}{T} \right),
\end{equation}
where $a$, $b$, $c$, and $d$ are positive constants determined by $q$, $L$ and $\mu$, along the the global sensitivity and privacy parameters. 
As it is evident from \eqref{eq: tradeoff 1}, in presence of data heterogeneity $(\lambda > 0)$, increasing $E$ can improve convergence for larger $\lambda$. 

\subsubsection{Model Compression v.s. Others} Consider the model compressor used in quantized-SGD~\cite{alistarh2017qsgd}, such that for any $\mathbf{x}\in\mathbb{R}^d$, the $i$-th element is quantized as
\begin{equation}
\label{eq: QSGD}
    Q_i(\mathbf{x}) = \|\mathbf{x} \|\cdot \mathrm{sign}(\mathbf{x}_i)\cdot \vartheta_i(\mathbf{x}, s),
\end{equation}
where $\vartheta_i(\mathbf{x}, s)$ is a random variable taking on value $\frac{l+1}{s}$ with probability $\frac{|\mathbf{x}_i|}{\|\mathbf{x}\|}s - l$ and $\frac{l}{s}$ otherwise. Here, $s\in\mathbb{Z}^+$ is the quantization level and $l\in[0,s)$ is an integer such that $\frac{|\mathbf{x}_i|}{\|\mathbf{x}\|}\in\left[\frac{l}{s}, \frac{l+1}{s} \right)$. To transmit $Q(\mathbf{x})$, three parts of information need to be encoded, including $\|\mathbf{x} \|^2$, $\left\{\mathrm{sign}(\mathbf{x}_i)\right\}_{i=1}^d$, and $\left\{\vartheta_i(\mathbf{x}, s)\right\}_{i=1}^d$. Assuming we are using the simplest one-hot encoder, jointly encoding $\left\{\mathrm{sign}(\mathbf{x}_i)\right\}_{i=1}^d$ and $\left\{\vartheta_i(\mathbf{x}, s)\right\}_{i=1}^d$ will cost $d\log_2(2s+1)$ bits, while $\|\mathbf{x} \|$ costs $32$ bits. Assume $d\log_2(2s+1)\gg 32$, we have
$
    \beta \approx d\log_2(2s+1).
$
By taking $q = \frac{\sqrt{d}}{s}$, the quantization loss $q$ is approximately
$
    \mathcal{O}\left(\frac{2\sqrt{d}}{2^{\beta/d} - 1}\right).
$
Clearly, $q$ will decrease rapidly with $\beta$. Furthermore, as we can see in Theorem \ref{th: convergence}, there is no term with $q$ and $\lambda$ coexisting, indicating that the impact of $q$ is invariant to non-i.i.d. data.
On the other hand, $\frac{E}{M}\sim\mathcal{O}\left(\beta\right)$ and the impact of $E$ is magnified under data heterogeneity. Thus, we can choose a relatively small $\beta$ for a good trade-off between model compression and the other two communication-reduction approaches.

\subsubsection{Impact of Privacy Measure}
As mentioned before, the local updates are clipped such that the global sensitivity is bounded. Theorem \ref{th: convergence} suggests that the convergence can be severely delayed for large $G$. This is due to the fact that in \ref{as: noise eq}, the noise variance quadratically grows with the $L_2$ sensitivity. 

Increasing the subsampling ratio $\gamma$, determined by $Eb/n_k$, amplifies the noise variance required for desired privacy guarantee. For large $E$, the last terms in \eqref{eq: theorem 1},

\begin{equation*}
    \frac{4096dG^2b^2(1+q)\ln\left(\frac{1.25Eb}{|\mathcal{D}_i|\delta} \right)}{M|\mathcal{D}_i|^2\epsilon^2}\frac{E^3}{T}
\end{equation*}
dominates convergence. In this case, increasing $b$, which consequently increases $\gamma$, would deteriorate the performance, hence our decision to employ small subsampling ratios to achieve privacy amplification. On the other hand, for small $E$ increasing $\gamma$ may not be as detrimental, or can even be beneficial, as is the case with the non-privatized algorithm. Clearly, increasing, $\epsilon$ and $\delta$ improve the convergence rate, but is not desirable as we would be opting for a risky and insufficient privacy guarantee. 

\begin{figure*}[h]
    \centering
    \subfloat[]{
      \includegraphics[width=70mm]{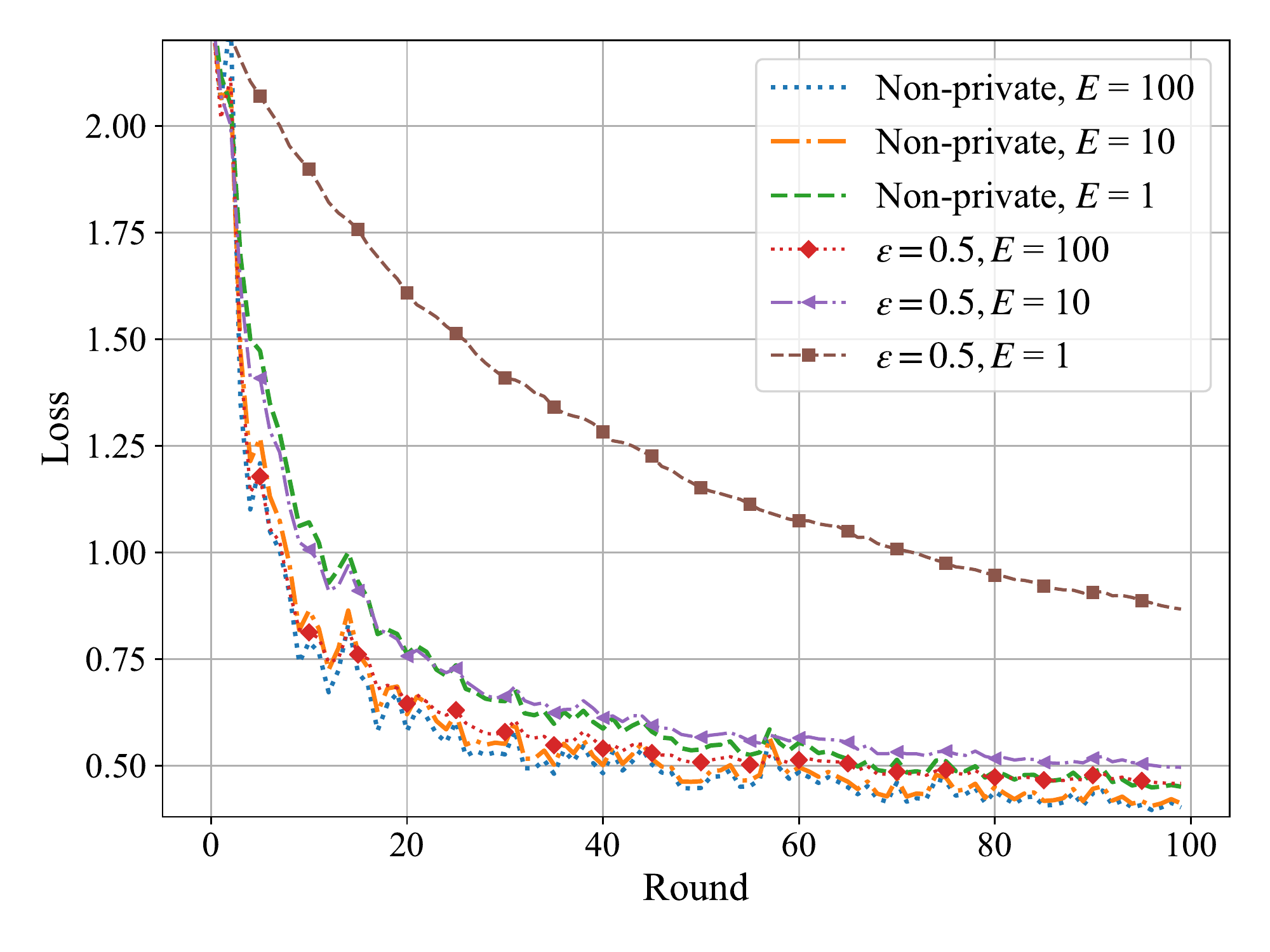}
    }
    \subfloat[]{
      \includegraphics[width=70mm]{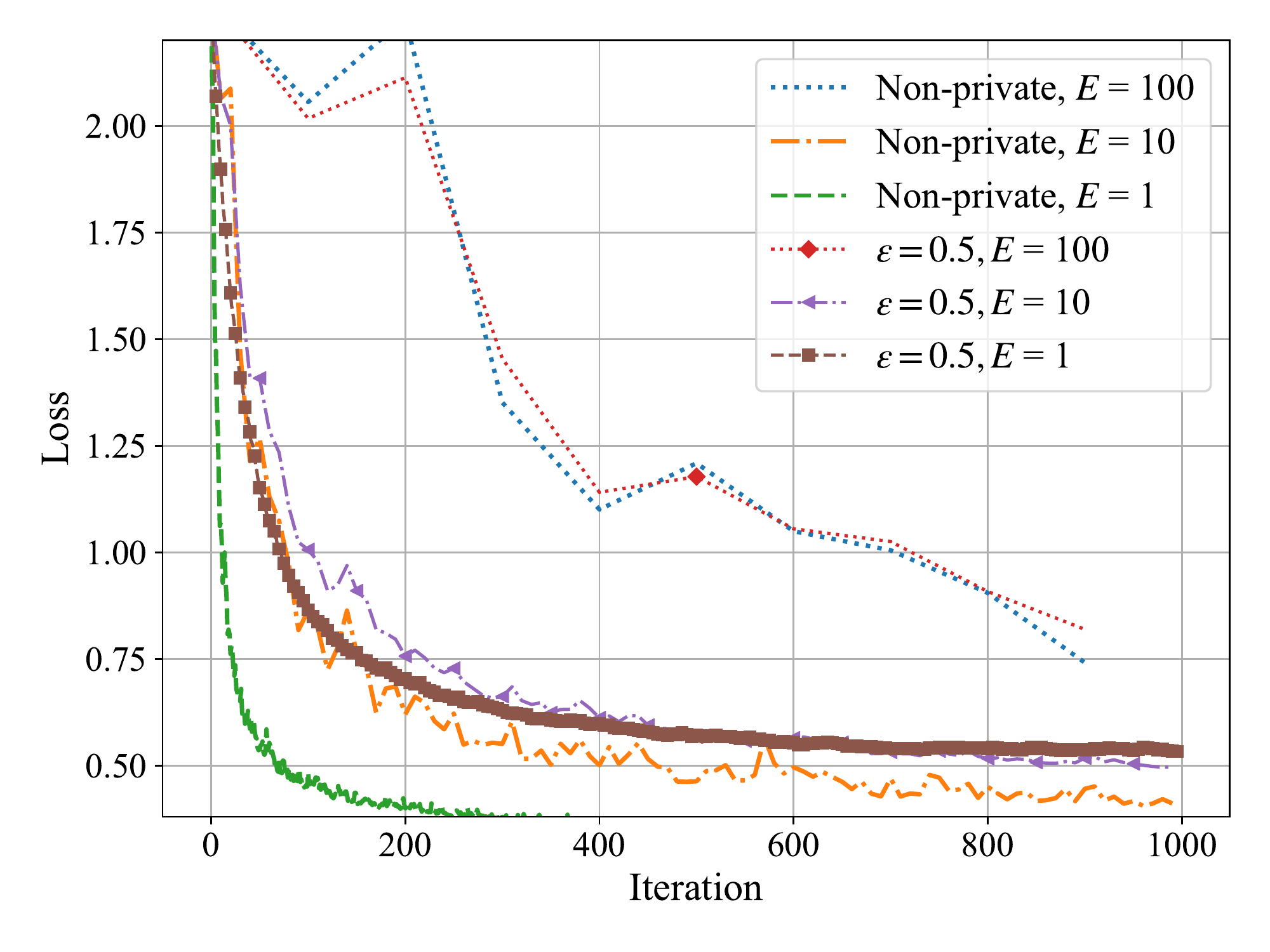}
    }
    \hspace{0mm}
    \subfloat[]{
      \includegraphics[width=70mm]{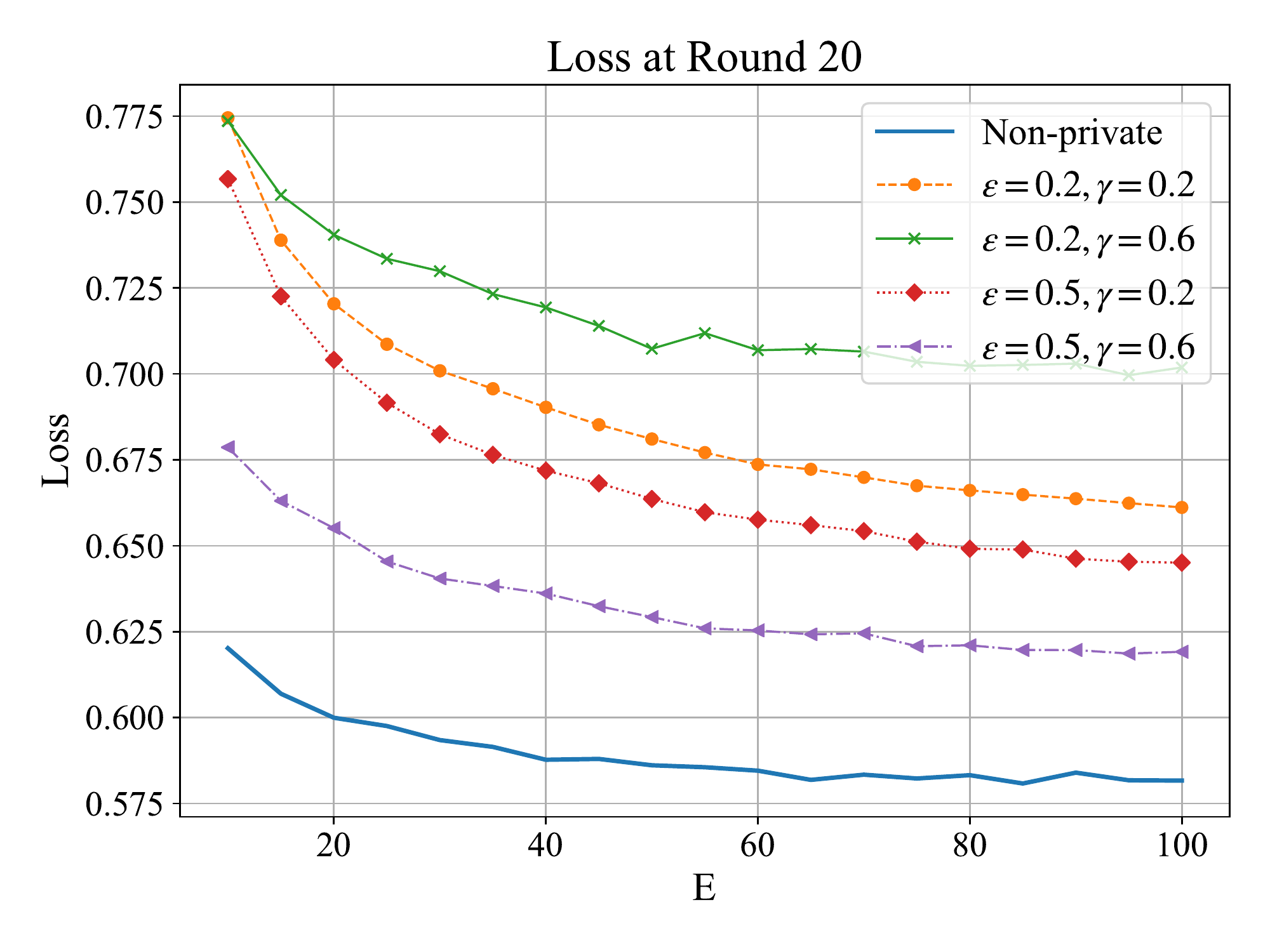}
    }
    \subfloat[]{
      \includegraphics[width=70mm]{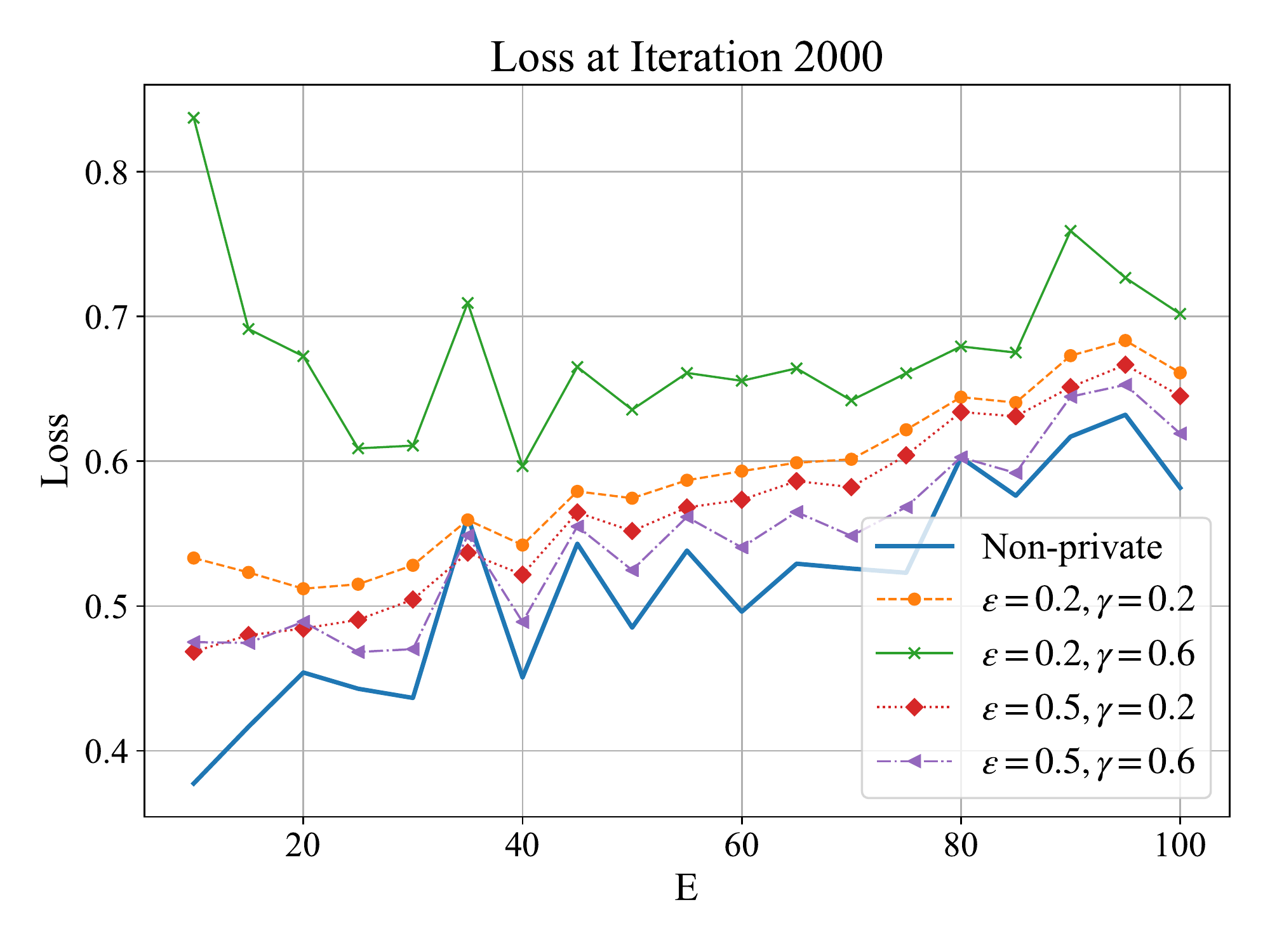}
    }
    \caption{The impact of $E$ on \texttt{HET\_MNIST}(2) dataset: (Top row): The training curves for different $E$ with $M=10$, $s=10$, $\gamma=0.2$ and $C=1$. (a) Over rounds (b) Over iterations. (Bottom row) The training loss (a) after $20$ communication rounds (b) after $2000$ iterations.}
    \label{fig: training_loss_E}
\end{figure*}

\begin{figure*}[h]
    \centering
    \subfloat[]{
      \includegraphics[width=70mm]{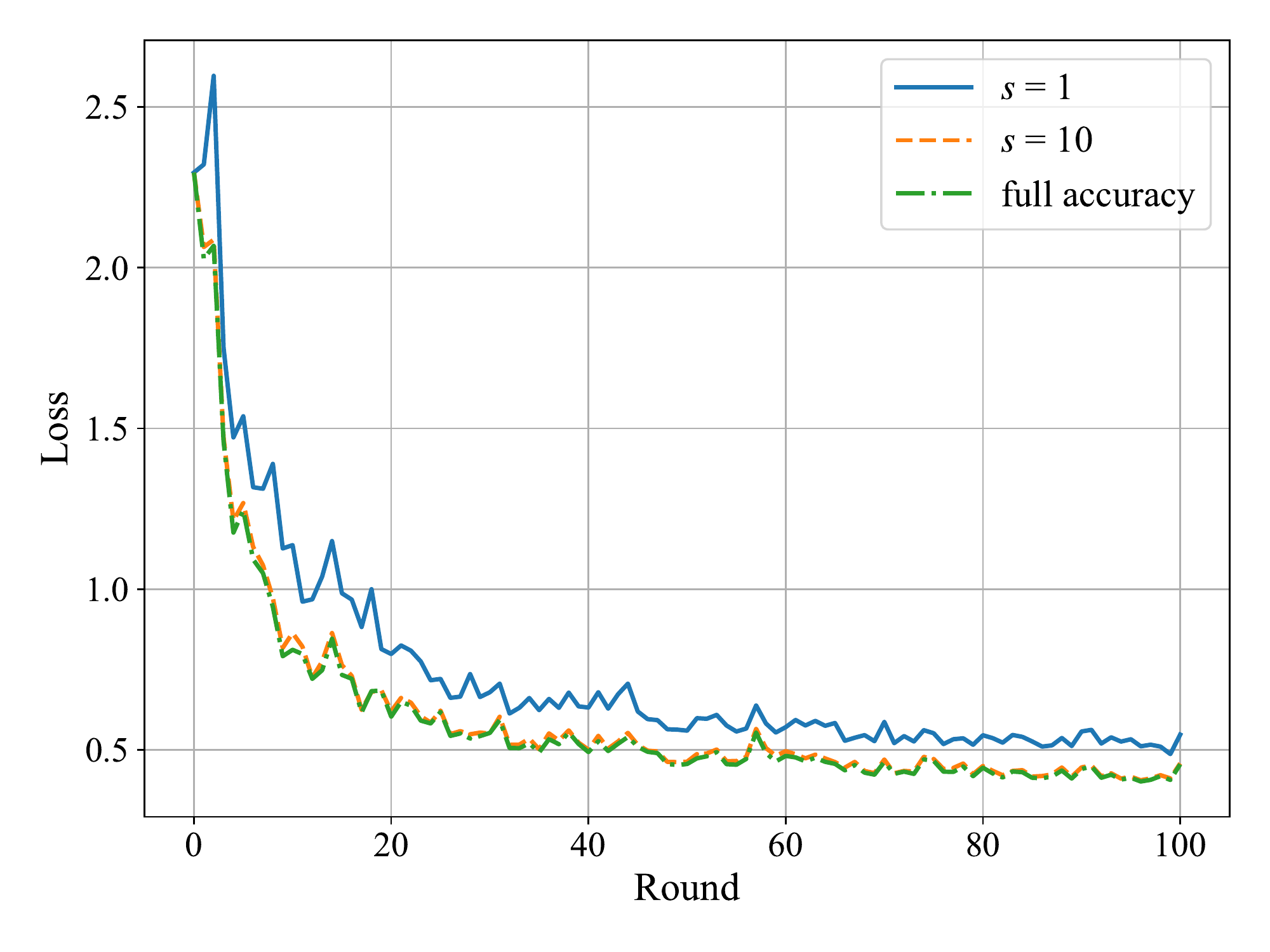}
    }
    \subfloat[]{
      \includegraphics[width=70mm]{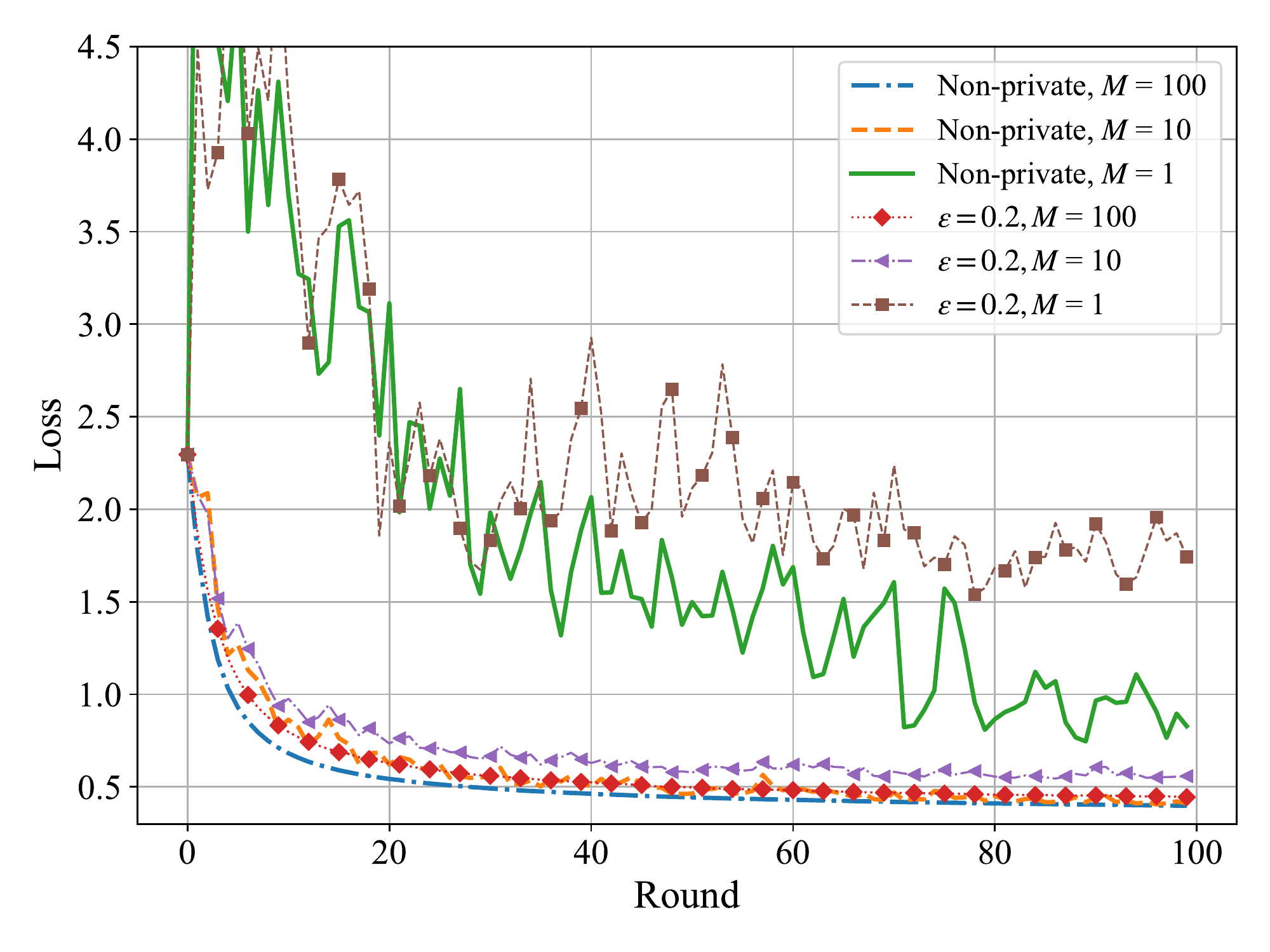}
    }
    \caption{Training curves for different $s$ and $M$ at each round on \texttt{HET\_MNIST}(2) dataset: (a) Different $s$ on the non-privatized model with $M=10$ and $E=10$. (b) Different $M$ with $E=10$, $s=10$, $\gamma=0.2$ and $C=1$.}
    \label{fig: training_loss_s_and_M}
\end{figure*}


\begin{figure*}[h]
    \centering
    \subfloat[]{
      \includegraphics[width=80mm]{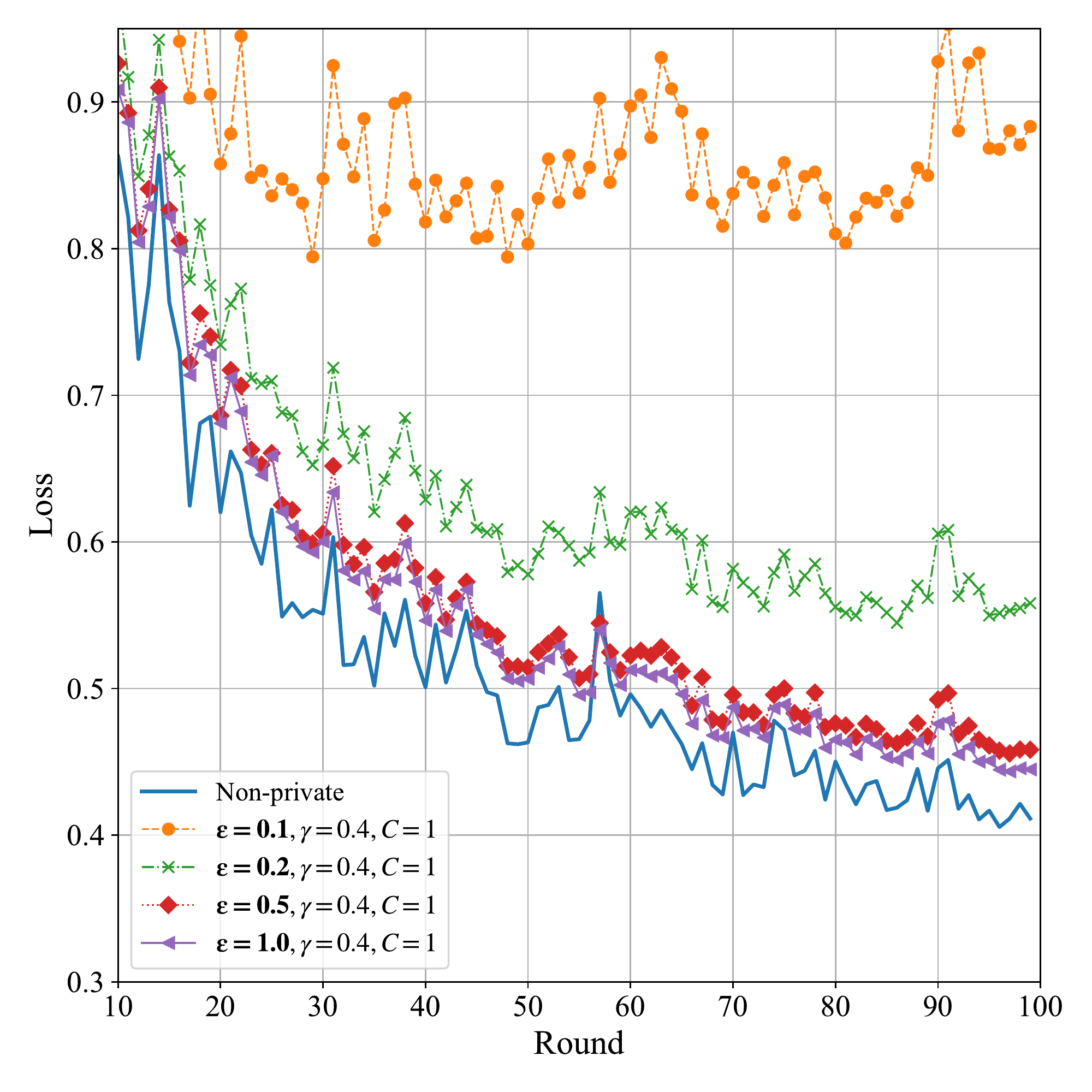}
    }
    \subfloat[]{
      \includegraphics[width=80mm]{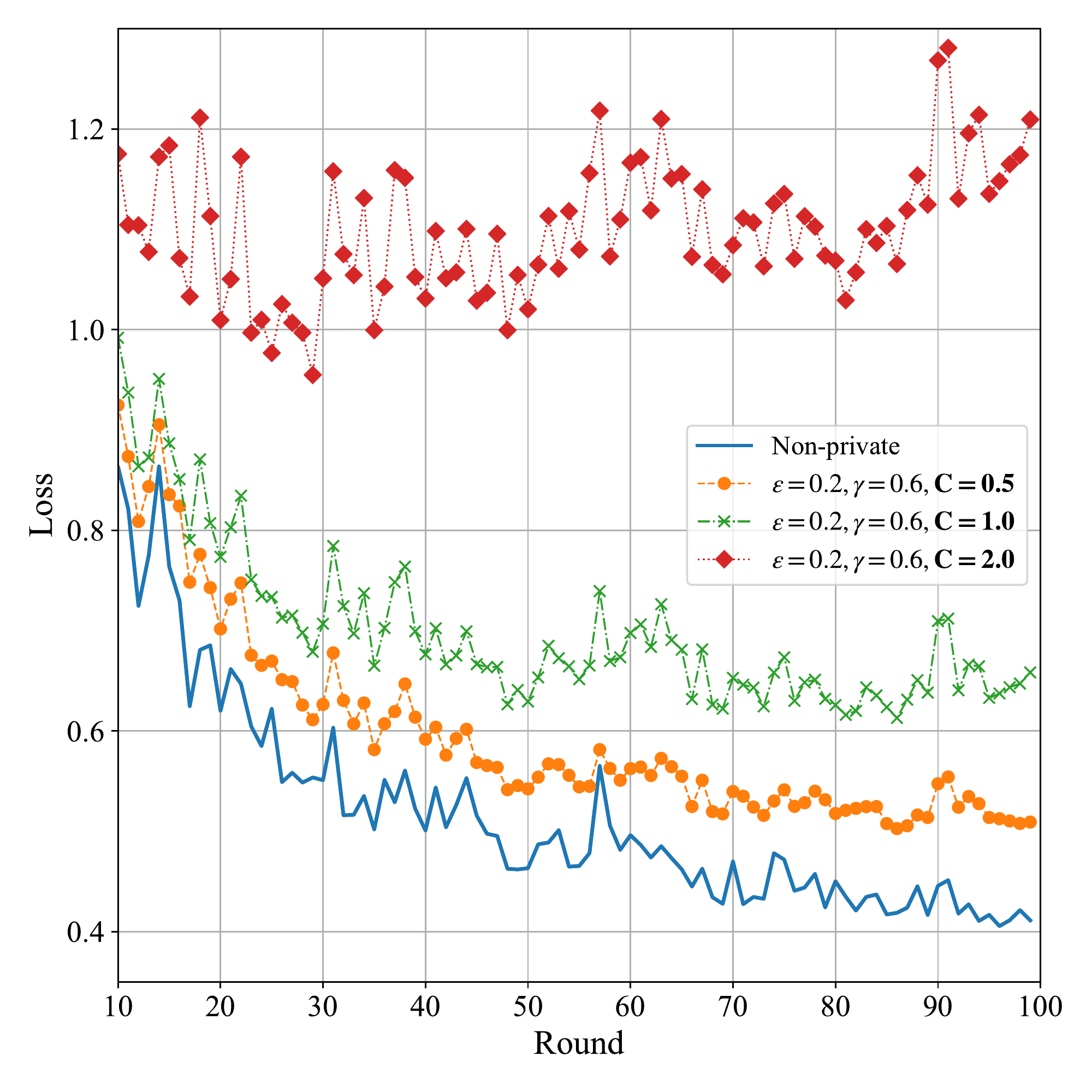}
    }
    \hspace{0mm}
    \subfloat[]{
      \includegraphics[width=80mm]{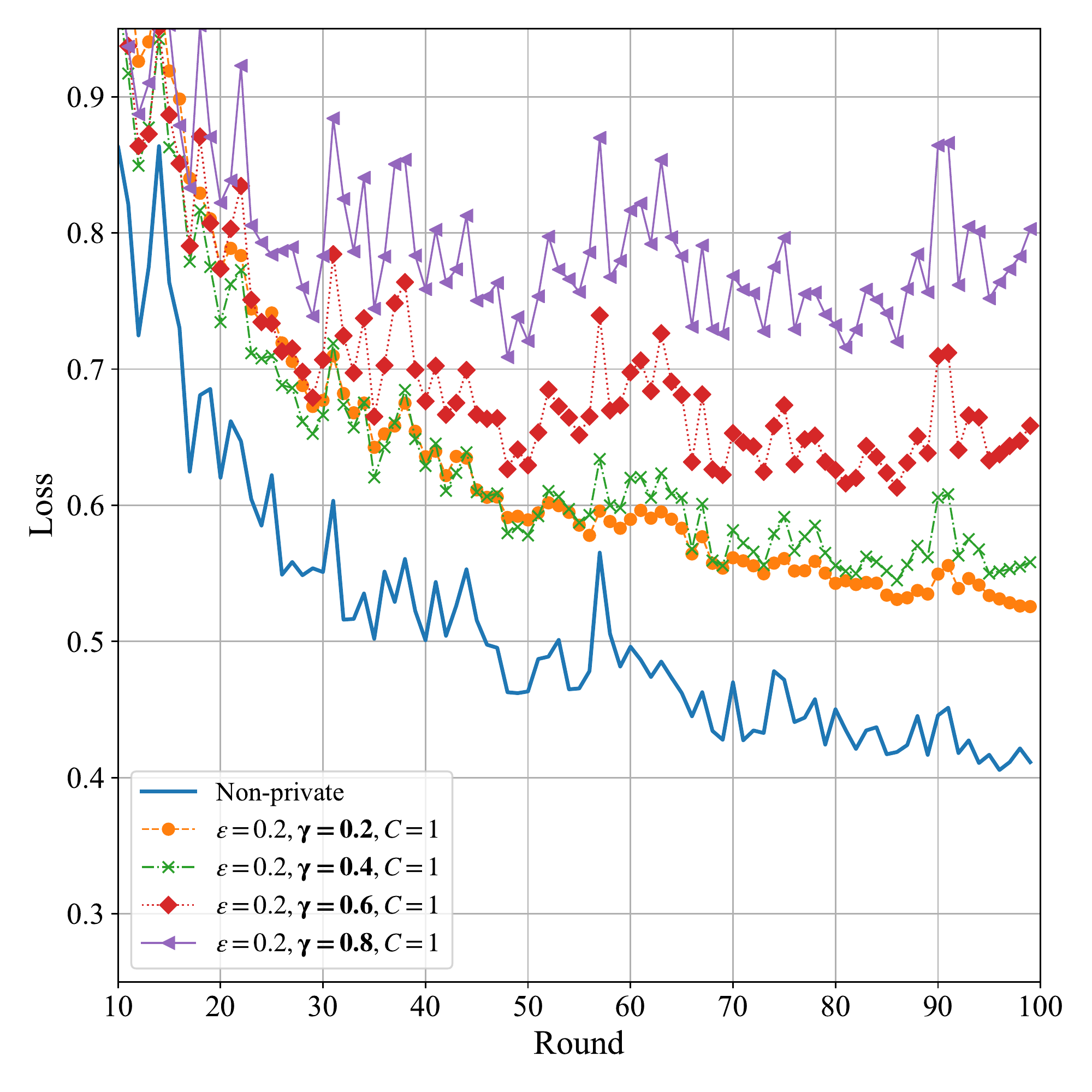}
    }
    \subfloat[]{
      \includegraphics[width=80mm]{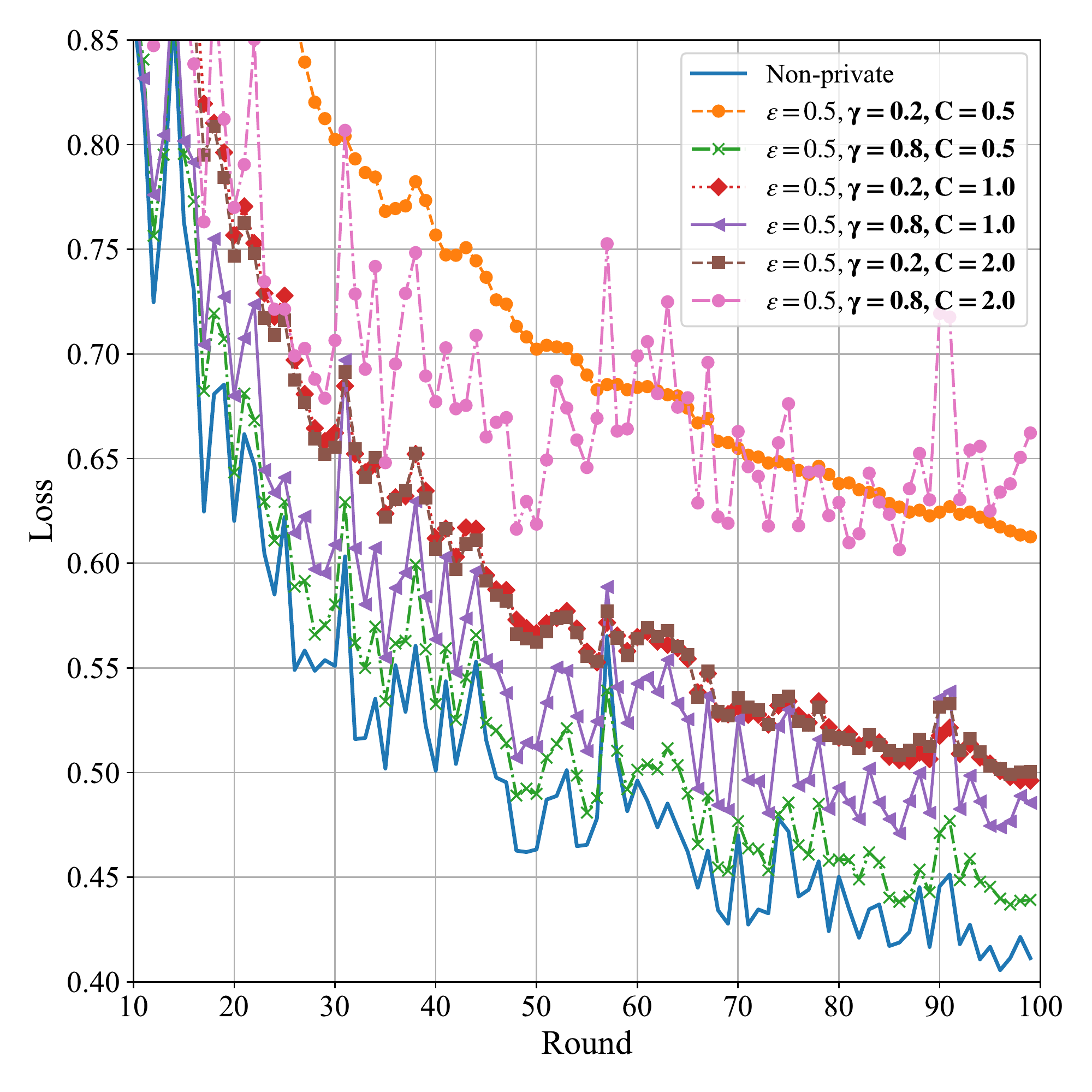}
    }
    \caption{The impact of different privacy-related parameters on convergence of the privatized FedPaq algorithm with $E=10$, $M=10$ and $s=10$ (a) for different $\varepsilon$ (b) Training curves of different $C$, (c) Training curves of different $gamma$, (c) Training curves for both $\gamma$ and $C$.}
    \label{fig: priv_loss}
\end{figure*}

\section{Performance Evaluation}

\subsubsection{Model and Dataset} 
The theoretical results are evaluated via a logistic regression model based on the widely-known MNIST dataset. The MNIST dataset is equally distributed among $N=100$ local devices. We control the degree of data heterogeneity by allowing a local device have access only to training samples for a fraction of all the $10$ digits. Consequently, we can generate ten datasets \texttt{HET\_MNIST(n\_digits)}, for $\texttt{n\_digits}=1,2,\cdots,10$. Clearly, choosing a smaller number of digits results in higher degree of data heterogeneity, while $\texttt{n\_digits}=10$ indicates no heterogeneity among the local datasets.

\subsubsection{Experiment Settings} The local models are aggregated for every $E$ iterations, such that there are $K = \left\lfloor \frac{T}{E} \right\rfloor$ communication rounds. Unless otherwise stated, $E=10$ local iterations are performed by each client, and number of communication rounds $K$ is fixed to $100$, with the total number of iterations $T$ accordingly calculated. The learning rate at the $k$-th round is set to be $\eta_k = \frac{\eta_0}{1 + kE/100}$, where $\eta_0=0.1$. The non-privatized setting refers to the vanilla FedPaq algorithm, with neither subsampling, clipping nor gradient perturbation. In the secure mode, a subset of local dataset with cardinality $\gamma n_i$ is randomly selected without replacement by each participating device $i\in\mathcal{S}_k$ at the start of each round to be used for local training. During training the per-sample gradients are clipped with respect to $C$ and proper noise is added prior to the transmission of local gradients.  Upon aggregation, the scheduled devices will send the (noisy) model updates compressed by the quantizer in \eqref{eq: QSGD} with quantization level $s$. We evaluate the impact of communication-reduction techniques on both the non-privatized and secure aggregated model over training rounds. Furthermore, we will also investigate how each of the parameters involved in privatizing the Secure FedPaq, ceteris paribus, influences the performance of the introduced federated learning scheme.

\subsubsection{Impact of Periodic Aggregation} 
Under the presence of data heterogeneity, the impact of $E$ is illustrated in Fig. \ref{fig: training_loss_E}. 
We observe that the accuracy of the model increases for larger $E$ per round, as illustrated in Fig. \ref{fig: training_loss_E} (a). This refers to the realistic case of constraining the number of communication rounds to a fixed number. Alternatively, Fig. \ref{fig: training_loss_E} (b) shows that fixing the number of global iterations $T$, larger $E$ in fact delays the convergence of the model at each iteration. The empirical performance verifies our analytical result in Theorem \ref{th: convergence}, that for fixed $T$, increasing $E$ would deteriorate the result. Intuitively, this means fewer opportunities for aggregating the local updates. However, proportionally increasing both $E$ and $T$ clearly helps both versions of FedPaq, although a diminishing gain is achieved. Fig. \ref{fig: training_loss_E} (c) and (d) show the loss curves when the number of rounds or the number of global iterations are fixed, respectively. Fig. \ref{fig: training_loss_E} confirms our interpretation of Theorem \ref{th: convergence} that larger subsampling ratio $\gamma$ would be more detrimental for larger values of $E$. 

\subsubsection{Impact of Partial Participation} 
\looseness=-1
As shown in Fig. \ref{fig: training_loss_s_and_M}(b), when $M$ is set to a small value, for example $M=1$, the training becomes extremely unstable during the first several training rounds. Note that the impact is even more pronounced for the privatized algorithm, consistent with the last term of (\ref{eq: theorem 1}) where the impact of perturbation is inhibited for larger $M$.


\subsubsection{Impact of Model Compression}
In Fig. \ref{fig: training_loss_s_and_M} (a), the loss curve for $s=10$ is almost overlapped with the one without quantization. Even for $s=1$, the performance degradation is small and the training is pretty stable. This motivates us to use lossy model compressors to save communication resources for smaller $E$ and larger $M$, which have significant impacts on the convergence of the global model. The same impact can be observed for the privatized setting, which is omitted to avoid polluting the figure. 

\subsubsection{Impact of Privacy Budget $\varepsilon$}
Increasing the privacy parameters $\varepsilon$ and $\delta$ result in additive noise of lower magnitude, which brings more accurate estimation while putting the privacy of clients at risk. We have set $\delta=10^{-4}$ across all the experiments. Fig. \ref{fig: priv_loss} (a) depicts the impact of $\varepsilon$ on performance of the mode. 
For $\varepsilon=0.1$, although a high level of privacy guarantee is ensured, the utility significantly declines and convergence may be hindered. 

\subsubsection{Impact of Gradient Clipping}
The global sensitivity $G$ appears in the numerator of multiple terms in Eq. \ref{eq: theorem 1} suggesting its major impact on the convergence rate of the algorithm. Fig. \ref{fig: priv_loss} (b) shows that larger $C$, inducing more noise, delays convergence of the privatized algorithm. In fact, one might claim that a smaller $C$ for clipping the gradients may be preferred to choosing larger $\varepsilon$ for achieving faster convergence. However, based on Fig. \ref{fig: priv_loss} (d), this is not always true.

\subsubsection{Impact of Subsampling}
As a privacy amplification measure, we are using subsampling to reduce the amount of noise that needs to be added to maintain a certain level of privacy guarantee, thus increasing the utility. Fig. \ref{fig: training_loss_E} (c) and (d) show how a larger subsampling ratio $\gamma$ has a more pronounced deteriorating impact for greater values of $E$, verifying our analytical results. Moreover, Fig. \ref{fig: priv_loss} (c) depicts the training loss of a few privatized models for various values of $\gamma$, but otherwise identical. Fig. \ref{fig: priv_loss} (d) show how $\gamma$ and $C$ jointly affect the performance of the model. One can observe that a smaller clipping of $C=0.5$ achieves the best or one of the worst privatized setting results based on which value of $\gamma$ is chosen. Although small $C$ and $\gamma$ lead to relatively negligible added noise, however, clipping the gradients acquired on such small subsample would significantly deteriorate convergence, as would be the case with non-privatized SGD. A more moderate clipping, namely $C=1$, would be more resilient in face potentially undesirable consequences of subsampling.

The theoretical and experimental results suggest some important design intuitions for federated learning systems: (1) The choice of model compression accuracy has little impacts on the convergence regardless of heterogeneity of the problem. This encourages the wide-spread use of low-accuracy quantizers to save the underlying communication overhead even in non-iid settings. (2) Although larger $E$ naturally incurs more computational cost for the clients, we realized that in non-iid setting, it can actually enhance the performance for fixed number of communication rounds $K$. (3) A smaller $M$ will reduce the convergence rate of the model. This behavior is even more noticeable for the privatized setting. (4) Subsampling becomes more important for larger $E$. (5) Generally, lower sampling ratio can achieve better convergence, but it may make the optimization unstable if the global sensitivity is very low.  


\section{Conclusion}
In this paper, we provided a comprehensive convergence analysis for a privacy augmented federated learning system by considering data heterogeneity as well as the communication constraints.
To be specific, three communication-reduction strategies namely model compression, partial device participation, and periodic aggregation are jointly considered under the capacity limit of the underlying MAC for model aggregation. Due to insufficiency of FL in maintaining the privacy of clients, a privacy measure based on deferential privacy is considered, introducing the privacy-augmented FedPaq algorithm. 
The impacts of differential privacy, data heterogeneity and the communication-reduction strategies were evaluated both theoretically and numerically. 
Our analysis provides important design intuitions for real-world federated learning systems that are limited by the communication capacity constraints in wireless networks. 


\bibliography{IEEEabrv,./ref}

\end{document}